\documentclass[sigconf]{acmart}
\DeclareMathOperator*{\argmin}{argmin}
\DeclareMathOperator*{\argmax}{arg\,max}
\usepackage{algorithm} 
\usepackage{algpseudocode}
\usepackage{mathtools}
\usepackage{amsmath}
\usepackage{soul}
\usepackage{wrapfig}
\usepackage{enumitem}
\usepackage{hyperref}
\usepackage{multirow}
\usepackage{xcolor}

\newtheoremstyle{sig}
  {}
  {}
  {\itshape}
  {}
  {\scshape}
  {.}
  {.5em}
  {#1 #2\thmnote{\quad(#3)}}

\theoremstyle{sig}
\newtheorem{theorem}{Theorem}[section]
\newtheorem{corollary}[theorem]{Corollary}
\newtheorem{assumption}{Assumption}

\newtheorem{lemma}[theorem]{Lemma}

\theoremstyle{definition}
\newtheorem{case}{Case}
\newtheorem{problem}{Problem}

\def\mathbi#1{\textbf{\em #1}}

\AtBeginDocument{%
  \providecommand\BibTeX{{%
    \normalfont B\kern-0.5em{\scshape i\kern-0.25em b}\kern-0.8em\TeX}}}

\setcopyright{acmcopyright}
\copyrightyear{2024}
\acmYear{2024}
\setcopyright{rightsretained}
\acmConference[KDD '24]{Proceedings of the 30th ACM SIGKDD Conference on Knowledge Discovery and Data Mining}{August 25--29, 2024}{Barcelona, Spain}
\acmBooktitle{Proceedings of the 30th ACM SIGKDD Conference on Knowledge Discovery and Data Mining (KDD '24), August 25--29, 2024, Barcelona, Spain}\acmDOI{10.1145/3637528.3672023}
\acmISBN{979-8-4007-0490-1/24/08}
\begin{document}

\title{Learning Flexible Time-windowed Granger Causality Integrating Heterogeneous Interventional Time Series Data}
\author{Ziyi Zhang}
\email{zyzhang@tamu.edu}
\affiliation{%
  \institution{Texas A\&M University}
  \city{College Station}
  \state{Texas}
  \country{USA}
}

\author{Shaogang Ren}
\email{shaogang@tamu.edu}
\affiliation{%
  \institution{Texas A\&M University}
  \city{College Station}
  \state{Texas}
  \country{USA}
}

\author{Xiaoning Qian}
\email{xqian@tamu.edu}
\affiliation{%
  \institution{Texas A\&M University, College Station, Texas}
  \institution{Brookhaven National Laboratory, Upton, New York}
  \country{USA}
}

\author{Nick Duffield}
\email{duffieldng@tamu.edu}
\affiliation{%
  \institution{Texas A\&M University}
  \city{College Station}
  \state{Texas}
  \country{USA}
}

\renewcommand{\shortauthors}{Zhang et al.}

\begin{abstract}
Granger causality, commonly used for inferring causal structures from time series data, has been adopted in widespread applications across various fields due to its intuitive explainability and high compatibility with emerging deep neural network prediction models. To alleviate challenges in better deciphering causal structures unambiguously from time series, the use of interventional data has become a practical approach. However, existing methods have yet to be explored in the context of imperfect interventions with unknown targets, which are more common and often more beneficial in a wide range of real-world applications. Additionally, the identifiability issues of Granger causality with unknown interventional targets in complex network models remain unsolved. Our work presents a theoretically-grounded method that infers Granger causal structure and identifies unknown targets by leveraging heterogeneous interventional time series data. We further illustrate that learning Granger causal structure and recovering interventional targets can mutually promote each other. Comparative experiments demonstrate that our method outperforms several robust baseline methods in learning Granger causal structure from interventional time series data.
\end{abstract}



\maketitle

\section{Introduction}
Time series data, capturing complex systems dynamic behaviors, are widely collected in many research areas, such as  economics, bio-informatics, and geo-informatics. Due to the rapid advancements in sensor and computing technologies, there has been a significant increase in research modeling time series data in recent years. Researchers have developed various methods leveraging time series data to perform related analysis such as optimization~\citep{ni2022policy, ni2022risk}, classification~\citep{ma2020adversarial, yang2021voice2series, younis2023flames2graph, zhang2023location}, clustering ~\citep{ma2019learning, zhang2021time, huijben2023som}, forecasting~\citep{vijay2023tsmixer, zhang2024towards}, and causal structure learning~\citep{khanna2019economy, marcinkevivcs2020interpretable, pamfil2020dynotears, tank2021neural, gao2022idyno, liu2023causal}. Among these tasks, causal structure learning is particularly challenging but important. Multivariate time series data, which capture the evolving states of multiple variables over time, facilitate deriving better systems understanding across various domains. Causal structure learning in multivariate time series data focuses on understanding how different variables influence each other. This knowledge is beneficial for explaining the data generation process and guiding the design of time series analysis methods~\cite{gong2023causal}.

Granger causality has been widely used for analyzing time series data to discover causal relationships in numerous real-world applications, including modern healthcare systems~\citep{wei2023granger}, medical time series generation~\cite{li2023causal} and time series anomaly detection~\cite{qiu2012granger}. Many methods for learning causal structures in time series have been developed based on the principles of Granger causality~\citep{xu2019scalable,khanna2019economy,chu2020inductive, marcinkevivcs2020interpretable,tank2021neural}. However, Granger causality tests based on linear models can be ineffective when faced with even slight non-linear causal relationships in the measurements. Consequently, a significant amount of research efforts have been focused on addressing issues for Granger causality considering non-linearities~\citep{khanna2019economy,tank2021neural,marcinkevivcs2020interpretable}.

\begin{figure}[ht]
  \centering
  \includegraphics[width=0.95\linewidth]{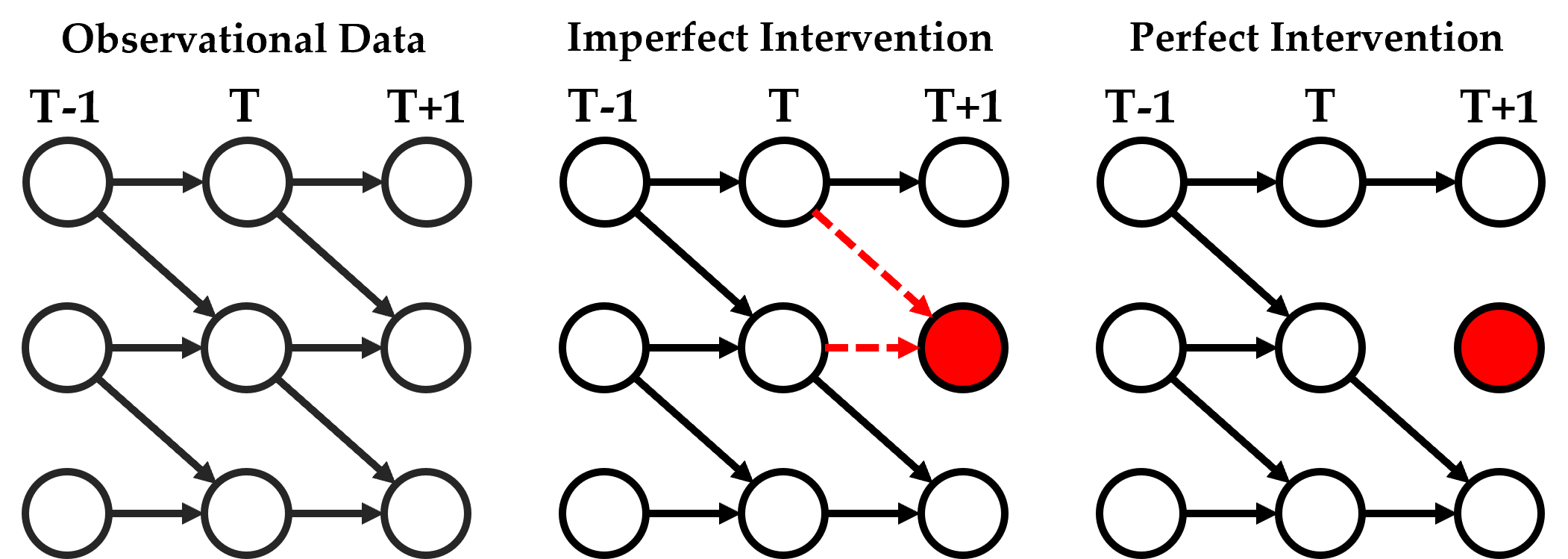}
  \captionsetup{width=\linewidth}
  \caption{Intervention types on time series: With known interventional target (red nodes), altered all causal relationships from parent nodes in imperfect interventions (red dotted lines) versus disconnection from parent nodes in perfect interventions.}
  \Description{}
  \label{fig:interv}
\end{figure}

Learning causal structures based solely on observational data is challenging~\cite{ren2021causal,ren2022flow} because, under the faithfulness assumption, the true causal structure can only be identified within a Markov Equivalence Class (MEC)~\cite{verma2022equivalence}. However, this identifiability improves when we consider interventional data. We have observed that domain experts might be able to gather interventional data in practice, where the underlying generative process varies across different conditions. This characteristic of distribution shift presents unique challenges as well as opportunities for learning causal structures in time series. In these scenarios, the causal structure can be identified within an Interventional Markov Equivalence Class ($\mathcal{I}$-MEC), which is a more specific subset of the Markov equivalence class~\citep{hauser2012characterization, yang2018characterizing, brouillard2020differentiable}. With sufficient interventional observations, the causal structure can be precisely identified~\citep{eberhardt2005number,eberhardt2008almost}. Numerous methods have approached causal structure learning with interventional data by framing it within a continuous optimization framework~\citep{brouillard2020differentiable,  pamfil2020dynotears, gao2022idyno}, incorporating a continuous acyclicity constraint~\cite{zheng2018dags}. To address identifiability challenges with time series data, the authors of~\cite{gao2022idyno} have extended the work of~\cite{brouillard2020differentiable, pamfil2020dynotears}, to 
effectively handle observational and interventional time series data under both perfect~\citep{eberhardt2007interventions,yang2018characterizing} and imperfect interventions \cite{peters2017elements} with known interventional targets~\cite{brouillard2020differentiable} (See Figure \ref{fig:interv}). However, in real-world applications, imperfect interventions with unknown targets are more common~\cite{zimmer2019loss}, requiring information about interventional targets limits their applications in more general cases. Therefore, learning causal structures from interventional time series data is still an open problem.

\begin{figure}[ht]
  \centering
  \includegraphics[width=0.85\linewidth]{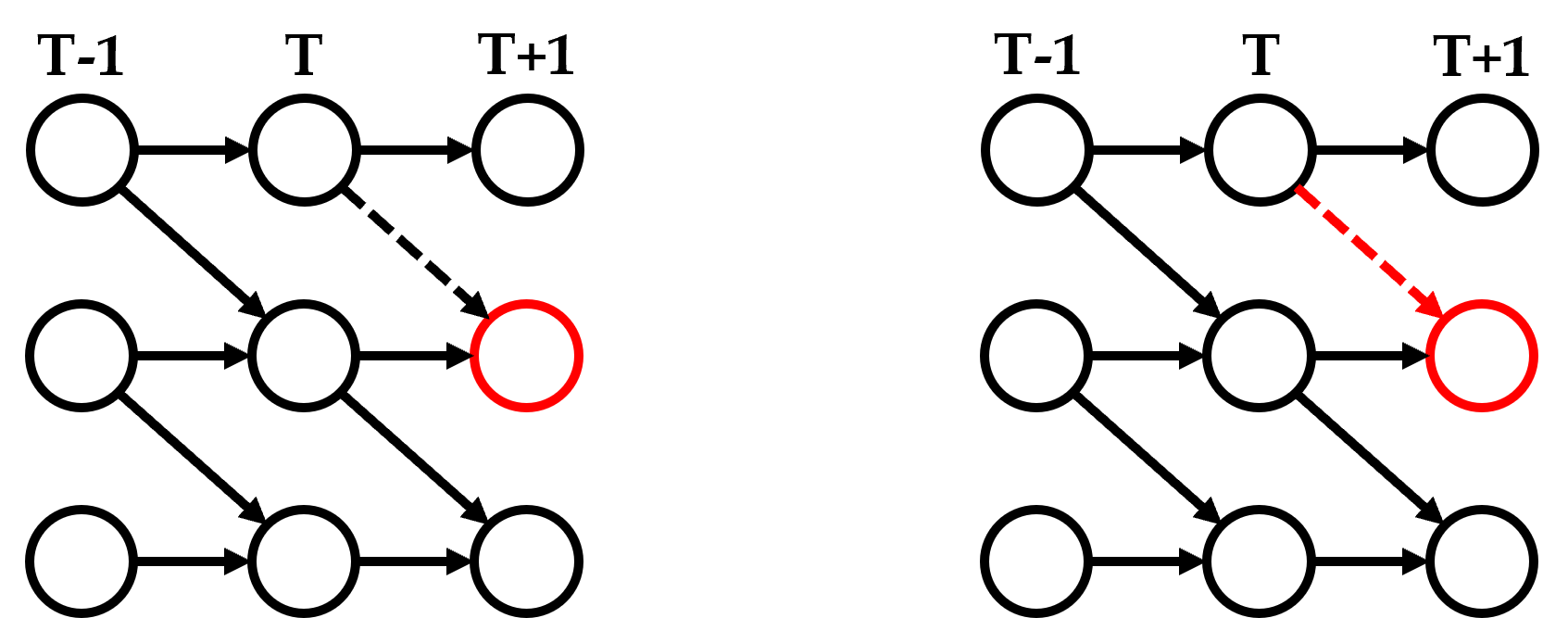}
  \captionsetup{width=\linewidth}
  \caption{(Left): Existing methods~(node-level imperfect intervention on an unknown target) can only identify the exact node(s); (Right): whereas our method~(edge-level intervention identification) can identify both the node(s) and exact edge(s).}
  \Description{}
  \label{fig:nae}
\end{figure}

\noindent In this paper, we emphasize on the following parts, compared to previous work:
\begin{itemize}[leftmargin=*]
    \item \textbf{Practicality.} Most methods require knowledge regarding interventional targets. However, in practical scenarios, distinguishing which variables originate from the non-intervened domain and identifying the exact interventional targets often proves to be challenging.
    \item \textbf{Accuracy.} In previous research, understanding of imperfect interventions has been limited to the node level. However, as illustrated in Figure \ref{fig:nae}, edge-level imperfect interventions can clarify the specific situations leading to imperfect interventions, which has not been explicitly studied.
    \item \textbf{Identifiability.} Despite the development of advanced Score-based~\citep{pamfil2020dynotears, gao2022idyno, liu2023causal} or Granger causality-based~\citep{xu2019scalable,khanna2019economy,chu2020inductive, marcinkevivcs2020interpretable,tank2021neural} causal structure learning methods for time series, issues related to the identifiability of Granger causality with unknown interventional targets remain unresolved.
\end{itemize}
Consequently, we introduce a theoretically-guaranteed Interventional Granger Causal structure learning (\textit{IGC}) method. This approach is designed for the simultaneous inference of Granger causal structure and the identification of unknown interventional targets at the edge level. It also leverages interventional time series data across multiple domains, efficiently differentiating among those that have not been intervened upon and those that have, especially in scenarios where interventional targets are unknown and the distinctions are not readily apparent. In summary, the main contributions of the paper are highlighted as:
\begin{itemize}[leftmargin=*]
    \item We have formalized the task of learning Granger causal structure from heterogeneous interventional time series data. The interventional targets are unknown and samples from observational distribution may be indistinguishable  from other interventional distribution.  
    \item A theoretically-guaranteed method called Interventional Granger Causal structure learning (\textit{IGC}) is developed to simultaneously infer Granger causal structure and identify unknown interventional targets at the edge level.
    \item We have shown that the exact minimization of the proposed objective will identify the $(\mathcal{I},\mathcal{D})$-Markov equivalence class of the ground truth graph in the context of unknown target setting, then resolve the identifiability issues of Granger causality.
    \item Extensive experiments on both synthetic and real-world time series data have demonstrated our proposed \emph{IGC} outperforms several robust baselines by utilizing interventional data.
\end{itemize}

\section{Related Work}
\textbf{Granger Causal Structure Learning:}
Much work has been conducted on inferring causal structure based on Granger causality in multivariate time series. Recent approaches for inferring Granger causal structure leverage the expressive power of neural network and are often based on regularized autoregressive models. \cite{arnold2007temporal} proposed the Lasso Granger method. \cite{tank2021neural} proposed the sparse-input multi-layer perceptron (MLP) and long short-term memory (LSTM) to model the nonlinear Granger causality within multivariate time series. \cite{khanna2019economy} integrated an efficient economy statistical recurrent unit architecture with input layer wights regularized in a group-wise manner. \cite{marcinkevivcs2020interpretable} proposed a generalized vector autoregression model that utilizes self-explaining neural networks (SENNs) for inferring Granger causal structure, with an additional focus on detecting signs of Granger-causal effects. \cite{cheng2023cuts,cheng2024cuts+} proposed a Granger causal discovery algorithm that builds a causal adjacency matrix for imputed and high-dimensional data using sparse regularization. Although these methods are powerful techniques for inferring Granger causal structure, they do not fully utilize interventional data, nor do they address the identifiability problem.\\
\textbf{Causal Structure Learning from Interventional Data:}
A series of parametric studies treat data from different distributions, often referred to as domains or environments, as interventional data. \cite{ghassami2018multi} studied the problem of causal structure learning in linear systems from observational data given in multiple domains, across which the causal coefficients may vary. \cite{yang2018characterizing} studied the problem of causal structure learning in the setting where both observational and interventional data is available and extended the identifiability results from perfect intervention \cite{hauser2012characterization} to general interventions. \cite{brouillard2020differentiable} proposed a differentiable causal structure learning method for static data that can leverage perfect, imperfect and unknown target interventions using score function to identify the $\mathcal{I}$-MEC. \cite{liu2023causal} propose a novel latent intervened non-stationary learning method to recover the domain indexes and the causal structure. \cite{gao2022idyno} extends \citep{brouillard2020differentiable, pamfil2020dynotears} to address both observational and interventional time series data, including perfect and imperfect interventions with known targets. However, effectively handling both observational and interventional time series data in an imperfect setting with unknown interventional targets remains a challenge.

\section{Preliminaries}
\noindent \textbf{Non-linear Granger Causality:} Consider multivariate time series $\mathcal{T}=\{\textbf{x}_1,...,\textbf{x}_T\}$, where $\textbf{x} \in \mathbb{R}^{d}$. Assume that causal relationships between variables are given by the following structural model:
\begin{equation}\label{eq1}
    x_{t+1}^{i} = g_i(x^{1}_{1:t},\ldots,x^{d}_{1:t}) \hspace{0.2cm} \text{for} \hspace{0.1cm} 1 \leq i \leq d,
\end{equation}
where $g_i(\cdot)$ is a function that specifies how the past values are mapped to series $i$. Time series $j$ is \emph{Granger non-causal} for time series $i$ if for all $x^{1}_{1:t},\ldots,x^{d}_{1:t}$ and all $x_{1:t}^{j} \neq \hat{x}_{1:t}^{j}$ \cite{tank2021neural}:
\begin{equation}\label{eq2}
    g_i(x^{1}_{1:t},\ldots,x^{j}_{1:t},\ldots,x^{d}_{1:t}) = g_i(x^{1}_{1:t},\ldots,\hat{x}^{j}_{1:t},\ldots,x^{d}_{1:t}).
\end{equation}

\noindent\textbf{Interventions:}
In the context of causal structure learning, an intervention on a variable $x_i$, involves altering its conditional probability $\textbf{P}(x_i|\textbf{PA}(x_i))$ to a new conditional probability $\widetilde{\textbf{P}}(x_i|\textbf{PA}(x_i))$, where $\textbf{PA}(x_i)$ is the set of parents of the node $x_i$ in the causal graph. It is possible to apply interventions to several variables at once. The set of variables on which $k$-th  interventions are made is referred to as the \textit{interventional targets}, symbolized by $\textbf{I}_k \in \mathbb{R}^{d}$. The \textit{interventional family} is defined as $\mathcal{I} := (\textbf{I}_1,\ldots,\textbf{I}_n)$, where $n$ represents the total number of interventions conducted.

\noindent\textbf{Types of Interventions:} The type of interventions depicted in Figure \ref{fig:interv} are generally categorized as imperfect intervention (also known as soft or parametric intervention)~\citep{peters2017elements, eaton2007exact}. In contrast, a specific case within this broad category is the perfect intervention (also referred to as hard or structural intervention), where $\textbf{P}(x_i|\textbf{PA}(x_i)) = \textbf{P}(x_i)$ \citep{eberhardt2007interventions,korb2004varieties, yang2018characterizing, brouillard2020differentiable}.

\section{Interventional Granger Causal Structure Learning}
In this section, we first discuss the challenge of learning Granger causal structure from interventional time series data in situations where the interventional targets are unknown, and samples from the observational distribution is indistinguishable from those of other interventional distributions. Subsequently, we propose our Interventional Granger Causal structure learning (\textit{IGC}) method to learn both the underlying Granger causal structure and unknown interventional targets across different environments.
\subsection{Granger Causality with Interventions}
First, we start with a linear Lasso Granger methodology capable of handling both observational and interventional data, drawing inspiration from the concepts applied to independent and identically distributed (i.i.d.) datasets \cite{brouillard2020differentiable} and structural vector autoregression model \cite{gao2022idyno}. The core principle of these methods involve constructing a Directed Acyclic Graph~(DAG) representing the ground truth causal graph from the interventional data. This is achieved by incorporating a distinct distribution family specifically for the intervened nodes within the log-likelihood objective. Unlike these methods that model the post-intervention distribution, our approach concentrates on comparing the distributions before and after the intervention under the framework of Granger causality to help interpret the impact of the intervention on time series data more clearly. Specifically, we employ $\textbf{W}_{e_0} \in \mathbb{R}^{d \times d}$ to represent the parameters of the density function for observational data under the condition of no interventions. For each intervention $\textbf{I}_k \in \mathcal{I}$, we define another corresponding set of parameters $\textbf{W}_{e_k} \in \mathbb{R}^{d \times d}$, which captures the differences in density functions before and after the $k$-th intervention. In other words, the density function after the $k$-th intervention can be represented as $\textbf{W}_{e_0} + \textbf{W}_{e_k}$. The collection of these parameters is denoted by $\mathbi{W} := \{\textbf{W}_{e_0},\textbf{W}_{e_1},...,\textbf{W}_{e_n}\}$. The integrated training loss function, as described in Equation (\ref{eq3}), takes into account both observational and interventional data:
\begin{equation}\label{eq3}
    \mathcal{L}(\textbf{X};\textbf{W}) = \sum_{k=1}^{n} \sum_{t=1}^{T} \sum_{\tau=1}^{l} \mathcal{L}_{k}(\textbf{X}_{t} - (\textbf{W}_{e_0} + \textbf{W}_{e_k})\textbf{X}_{t-\tau}),
\end{equation}
where $n$ represents the total number of interventions and $\mathcal{L}_k$ signifies the training loss based on the time series data from the $k$-th intervention. In this model, we do not know which environment among $n$ environments is non-intervend and we assume that $\textbf{W}_{e_0}$ remains constant across $n$ different environments, interventions, domains, or distributions. A time series $j$ is \emph{Granger non-causal} for time series $i$ if and only if the corresponding weight $\textbf{w}_{ij}$ in the matrix $\textbf{W}_{e_0}$ is zero. Intuitively,  if all elements in $\textbf{W}_{e_k}$ are zero, this suggests that the $k$-th environment is non-intervened. The optimization process can be expressed as follows:
\begin{equation}\label{eq4}
    \min_{\textbf{W}_{e_0},\textbf{W}_{e_k}}\sum_{k=1}^{n}\sum_{t=1}^{T} \sum_{\tau=1}^{l} ||\textbf{X}_{t} - (\textbf{W}_{e_0} + \textbf{W}_{e_k})\textbf{X}_{t-\tau}||_2^{2} + \lambda \Omega(\textbf{W}_{e_0},\textbf{W}_{e_k}),
\end{equation}
where $\tau$ is the time lag. The final estimated Granger causal structure without interventions is represented by $\textbf{W}_{e_0}$, and $\textbf{W}_{e_k}$ denotes the underlying intervention structures after the $k$-th intervention. The implementation details of the regularization penalty term $\Omega(\textbf{W}_{e_0},\textbf{W}_{e_k})$ will be discussed in the following section.

\subsection{Non-linear Granger Causality with Interventions}
\label{igc}
Linear Granger causal models, with their simplicity and straightforwardness, provide a clear but often oversimplified view of relationships among variables. In practice, it is challenging to model the highly non-linear relationships among multiple variables from time series data. In our proposed $\textit{IGC}$, we assume that there exist functions $f_{i}: \mathbb{R}^{d \times T} \rightarrow \mathbb{R}$ and $g_{i}: \mathbb{R}^{d \times T} \rightarrow \mathbb{R}$ such that:
\begin{equation}\label{eq5}
    \mathbb{E}[x_{i,t+1}|\textbf{PA}(x_{i,t+1})] = f_{i}(\textbf{X}_{t:t-T}) + g_{i}(\textbf{X}_{t:t-T}).
\end{equation}
$f_i(\textbf{x}_1,..,\textbf{x}_d)$ does not depend on $\textbf{x}_{k} \in \mathbb{R}^{T}$ if $\textbf{x}_k \cap \textbf{PA}(x_{i,t+1}) = \emptyset$; and $g_i(\textbf{X}_{t:t-T}) = 0$ if $x_{i,t+1}$ is not intervened, under the assumption of no instantaneous effects \cite{peters2013causal}. Thus, our objective is to learn $f = (f_1,...,f_d)$ and $g  = (g_1,...,g_d)$ such that the estimated Granger causal structure from $f$ and interventional targets from $g$. 

Let us first define $\overline{\textbf{W}}_{e_k} = \textbf{W}_{e_0} + \textbf{W}_{e_k}$ and concentrate on a single variable $x_i$ within a specific environment $e_k$. We define a set of parameters, which can be represented as: $\phi_{e_k}^{i}$, where $\phi^{i}_{e_k}=\{\overline{\textbf{W}}_{:,1,e_k}^{i},\ldots,\overline{\textbf{W}}_{:,d,e_k}^{i}\}$ and $\phi_{e_k}=\{\phi^{1}_{e_k},\ldots,\phi^{d}_{e_k}\}$. Then the overall objective becomes:
\begin{equation}\label{eq6}
\begin{aligned}
    \min_{\phi_{e_k}} \sum_{i=1}^{d} \sum^{T}_{t=2} ||x_{i,t}^{e_k} - \textbf{F}_{i}(x_{t-1:1}^{e_k};\mathcal{\phi}_{e_k}^{i})||_2^{2} + \lambda \sum_{i=1}^{d}\sum_{j=1}^{d} ||\overline{\textbf{W}}_{:,j,e_k}^{i}||_{2}, 
\end{aligned}
\end{equation}
where $\textbf{F}_{i}(\cdot)$ is defined as: 
\begin{equation}\label{eq7}
    \textbf{F}_{i}(x^{e_k}; \phi_{e_k}^{i}) = f_{i}(x^{e_k};\textbf{W}_{e_0}^{i}) + g_{i}(x^{e_k};\textbf{W}_{e_k}^{i}),
\end{equation}
and $\textbf{F}_i(\cdot)$ generates the estimate $\hat{x}_i$ for the next timestep in $e_k$, time series $j$ is \textit{Granger non-causal} for time series $i$ in the $e_k$ if and only if $\overline{\textbf{W}}_{:,j,e_k}^{i}$ is zero.
\noindent With the above proposed objective and heterogeneous interventional time series data from $n$ environments, we propose minimizing Equation (\ref{eq8}) to prioritize the discovery of the Granger causal structure that remains consistent across all environments $\mathcal{E} = \{e_1,\ldots,e_n\}$.
\begin{equation}\label{eq8}
\begin{aligned}
    \min_{\phi} & \scalebox{1.5}{$\sum$}_{k=1}^{n} \scalebox{1.5}{$\sum$}_{i=1}^{d} \scalebox{1.5}{$\sum$}^{T}_{t=2} || x_{i,t}^{e_k} - \textbf{F}_{i}(x_{t-1:1}^{e_k};\mathcal{\phi}_{e_k}^{i})||_2^{2}\\
    &+ \textstyle \lambda \scalebox{1.5}{$\sum$}_{i=1}^{d} \scalebox{1.5}{$\sum$}_{j=1}^{d} ||( \overline{\textbf{W}}_{:,j,e_{1}}^{i},...,\overline{\textbf{W}}_{:,j,e_{n}}^{i})||_{2}, 
\end{aligned}
\end{equation}
where $\phi = \{\phi_{e_1},\ldots,\phi_{e_n}\}$ represents the collection of parameters across all $n$ environments.

\noindent To learn the unknown interventional targets while maintaining consistency in the Granger causal structure, the overall penalized objective becomes:
\begin{equation}\label{eq9}
\begin{aligned}
    \min_{\phi} & \scalebox{1.5}{$\sum$}_{k=1}^{n} \scalebox{1.5}{$\sum$}_{t=2}^{T} \scalebox{1.5}{$\sum$}_{i=1}^{d} \scalebox{1.5}{$\sum$}_{j=1}^{d} || x_{i,t}^{e_k} - \textbf{F}_{ij}(x_{j,t-1:1}^{e_k}; \overline{\textbf{W}}_{:,j,e_{k}}^{i}) ||_2^2\\
    &+ (1-\alpha)\lambda \scalebox{1.5}{$\sum$}_{i=1}^{d} \scalebox{1.5}{$\sum$}_{j=1}^{d} ||(\textbf{W}_{:,j,e_{0}}^{i}, \textbf{W}_{:,j,e_{1}}^{i},...,\textbf{W}_{:,j,e_{n}}^{i})||_{2} \\
    &+ \alpha \lambda \textstyle \scalebox{1.5}{$\sum$}_{i=1}^{d} \scalebox{1.5}{$\sum$}_{j=1}^{d} \scalebox{1.5}{$\sum$}_{k=1}^{n}||\textbf{W}_{:,j,e_{k}}^{i}||_2, 
\end{aligned}
\end{equation}
where $\alpha \in (0,1)$ controls the tradeoff in sparsity across and within groups. After learning, time series $j$ is \textit{Granger non-causal} to $i$ if $\textbf{W}_{:,j,e_0}^{i}$ is zero across $k$ distributions. Furthermore, there is no intervention from time series $j$ to $i$ in the $k$-th distribution if $\textbf{W}_{:,j,e_k}^{i}$ is zero, which can be mathematically expressed as:
\begin{equation}
    \textbf{P}(x_{i,t}^{e_0}|x_{j,t-1:t-\tau}^{e_0}) = \textbf{P}(x_{i,t}^{e_k}|x_{j,t-1:t-\tau}^{e_k}).
\end{equation}

\subsection{Model Architecture}
\begin{figure}[ht]
  \centering
  \includegraphics[width=0.9\linewidth]{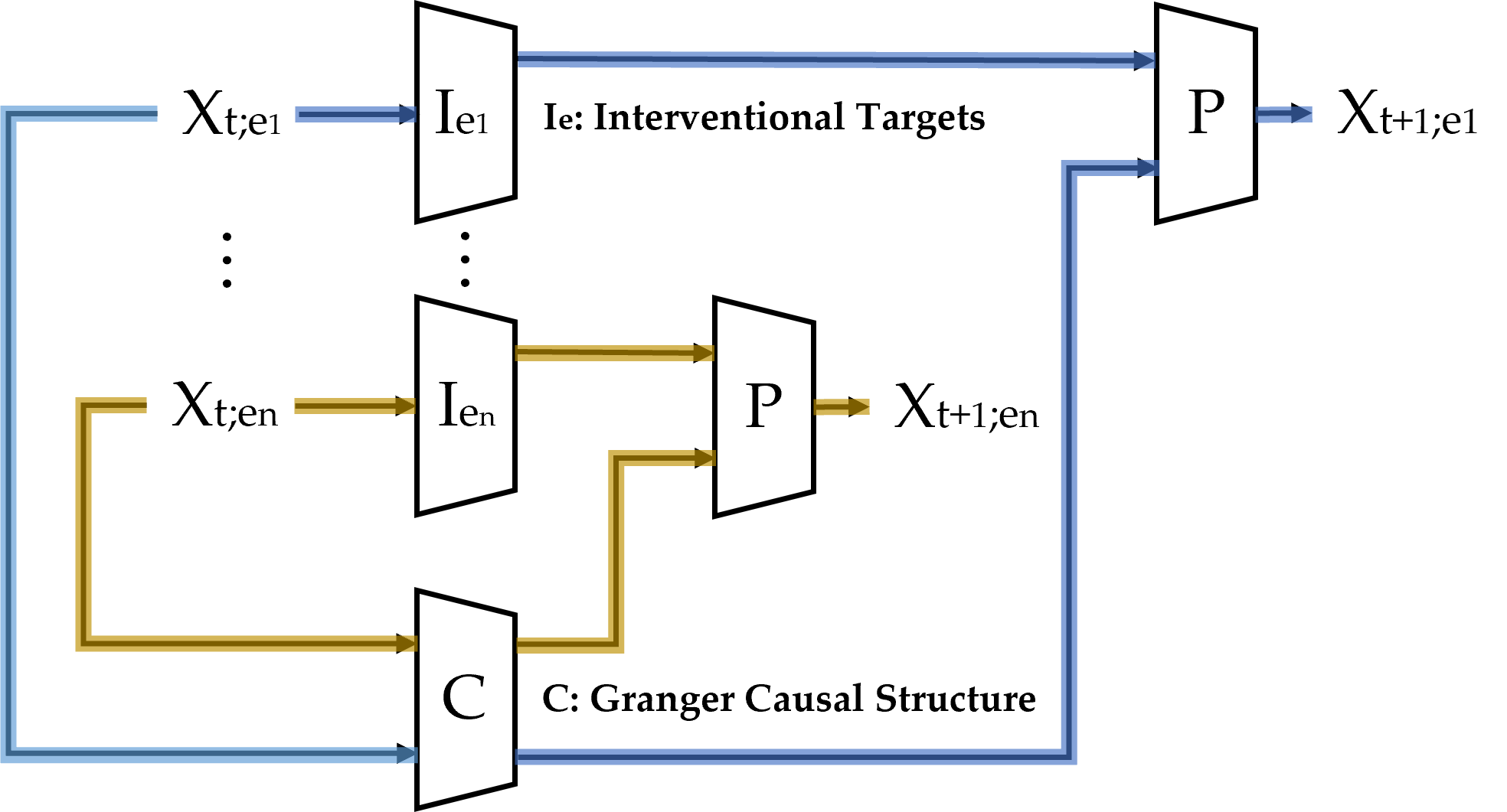}
  \caption{The information flow in various environments is represented by different colors. During the learning process, the prediction network (\textbf{P}) generates data for the next timestep. Information about unknown targets is contained within the intervention networks ($\textbf{I}_e$), and the Granger causal structure is captured within the causal network (\textbf{C}).}
  \label{fig:arch}
  \Description{}
\end{figure}

In line with the concepts presented in Section \ref{igc},  $\textit{IGC}$ utilizes historical time series data as its input and forecasts the data for the subsequent timestep as its output. The principal contribution of our study is the integration of heterogeneous interventional time series data, which aids in the identification of both the Granger causal structure and the interventional targets. The architecture of the model is illustrated in Figure \ref{fig:arch}.

\noindent \textbf{Intervention Networks.} Consider a set of time series data $\textbf{X} = \{\textbf{X}_{e_1},\ldots, \textbf{X}_{e_n} \}$ from $n$ environments or distributions. For each specific environment $e_k$, there exists an intervention network $\textbf{I}_{e_k} = \{\textbf{I}_{e_k}^{1},\ldots,\textbf{I}_{e_k}^{d}\}$, where each function $\textbf{I}_{e_k}^{i}$ is defined as:
\begin{equation}
    \textbf{I}_{e_k}^{i}(\textbf{X}_{t;e_k};\textbf{W}_{e_k}^{i}): \mathbb{R}^{T \times d} \rightarrow \mathbb{R}^{T \times h}.
\end{equation}
In this context, $\textbf{I}_{e_k}^{i}(\cdot)$ represents the intervention network for node $i$ in $e_k$,  $\textbf{X}_{t;e_k} \in \mathbb{R}^{T \times d}$ is the historical multivariate time series data in $e_k$, and $\textbf{W}_{e_k}^{i} \in \mathbb{R}^{d \times h}$ denotes the parameters of the intervention network $\textbf{I}_{e_k}^{i}$.

\noindent \textbf{Granger Causal Network.} For the time series data set $\textbf{X}$, a shared Granger causal network is applicable to all environments within $\textbf{X}$. This network is defined as $\textbf{C} = \{\textbf{C}_{1},\ldots, \textbf{C}_{d}\}$, where each $\textbf{C}_{i}$ is described by the function:
\begin{equation}
    \textbf{C}_{i}(\textbf{X}_{t};\textbf{W}_{e_0}^{i}): \mathbb{R}^{T \times d} \rightarrow \mathbb{R}^{T \times h}.
\end{equation}
In this context, $\textbf{X}_{t}$ represents the historical multivariate time series data from each environment within $\textbf{X}$ presented in sequence, and $\textbf{W}_{e_0}^{i} \in \mathbb{R}^{d \times h}$ are the parameters of the Granger causal network for node $i$.

\noindent \textbf{Information Aggregator.} After generating both the intervention information and the Granger causal information, we use a mechanism to aggregate them:
\begin{equation}
    \textbf{Z}_{t;e_k}^{i} = \textbf{Agg} (\textbf{I}_{e_k}^{i}(\textbf{X}_{t;e_k};\textbf{W}_{e_k}^{i}), \textbf{C}_{i}(\textbf{X}_{t;e_k};\textbf{W}_{e_0}^{i})), 
\end{equation}
where $\textbf{Agg}(\cdot)$ is an aggregation function and we have adopted summation in our experiments. We leave a learnable aggregation operator as a future research direction.

\noindent \textbf{Prediction Network.} The prediction network $\textbf{P}$ is designed to forecast the $i$-th data point for the subsequent timestep:
\begin{equation}
\widehat{\textbf{X}}_{t+1;e_k}^{i} = \textbf{P}(\textbf{Z}_{t;e_k}^{i}), 
\end{equation}
where $\widehat{\textbf{X}}_{t+1;e_k}^{i} \in \mathbb{R}$ represents the predicted value at timestep $t+1$, while $\textbf{Z}_{t;e_k}^{i} \in \mathbb{R}^{T \times h}$ denotes the aggregated embedding obtained from the previous step.

\noindent To enhance flexibility, components such as $\textbf{I}_e(\cdot)$, $\textbf{C}(\cdot)$, and $\textbf{P}(\cdot)$ can be effectively modeled using a variety of neural network architectures, including MLP, LSTM, SENNs, and Transformer. In our experiments, we employed MLPs and trained the model with respect to Equation (\ref{eq9}). As illustrated in Figure \ref{fig:arch}, information about unknown targets is obtained from the intervention networks $\textbf{I}_e$, while the Granger causal structure is estimated from the causal network \textbf{C}. The \emph{IGC} operates under the Assumption \ref{a1}.
\begin{assumption}\label{a1}
    (Causal Consistency). There exists a consistent causal structure and common parameters $\mathbf{W}_{e_0}$ across different environments. The dissimilarity between the parameters for one environment and the common parameters $\mathbf{W}_{e_0}$ lies within a range defined by a lower bound $\epsilon_l$, and an upper bound $\epsilon_u$. This range captures the extent of variation allowed between the common parameters and different environments. To avoid identical data across environments, the condition $\epsilon_l = 0$ indicates that there is no significant intervention. Mathematically it can be expressed as: $\epsilon_l \leq |\mathbf{W}_{e_k}| \leq \epsilon_u, \forall 1 \leq k \leq n, 0 \leq \epsilon_l \leq \epsilon_u$.
\end{assumption}
\subsection{Optimizing the Penalized Objective}
To optimize the objective stated in Equation (\ref{eq9}) for the proposed \emph{IGC} method, we use proximal gradient descent \cite{parikh2014proximal}, which is particularly beneficial for our purposes as it results exact zeros in the columns of input parameters, an essential aspect for interpreting Granger non-causality and intervention within our framework. The proximal operator is the group-wise soft-thresholding operator. Detailed updates of the proximal gradient descent are included in the Appendix \ref{pgd}. The proximal steps on the input weights for the penalty in Equation (\ref{eq9}) is shown in Algorithm \ref{alg:psp}, where $\textbf{Soft}(\cdot)$ is a group soft-thresholding operator on the input weights \cite{parikh2014proximal}.

\begin{algorithm}
\caption{Proximal steps for the penalty in Equation (\ref{eq9})}\label{alg:psp}
\begin{algorithmic}[1]
\Procedure{\textbf{Input}}{%
  $\alpha>0$, $\lambda>0$, $(\textbf{W}_{:,j,e_0}^{i},...,\textbf{W}_{:,j,e_k}^{i})$
}
\For{$k=1$ to $n$}
\State $\textbf{W}_{:,j,e_k}^{i} = \textbf{Soft}_{\alpha\lambda}(\textbf{W}_{:,j,e_k}^{i})$
\EndFor
\State $(\textbf{W}_{:,j,e_0}^{i},...,\textbf{W}_{:,j,e_k}^{i}) = \textbf{Soft}_{(1-\alpha)\lambda}((\textbf{W}_{:,j,e_0}^{i},...,\textbf{W}_{:,j,e_k}^{i}))$
\State \textbf{return} $(\textbf{W}_{:,j,e_0}^{i},...,\textbf{W}_{:,j,e_k}^{i})$
\EndProcedure
\end{algorithmic}
\end{algorithm}
\section{Identifiability}
\begin{figure}[ht]
  \centering
  \includegraphics[width=0.85\linewidth]{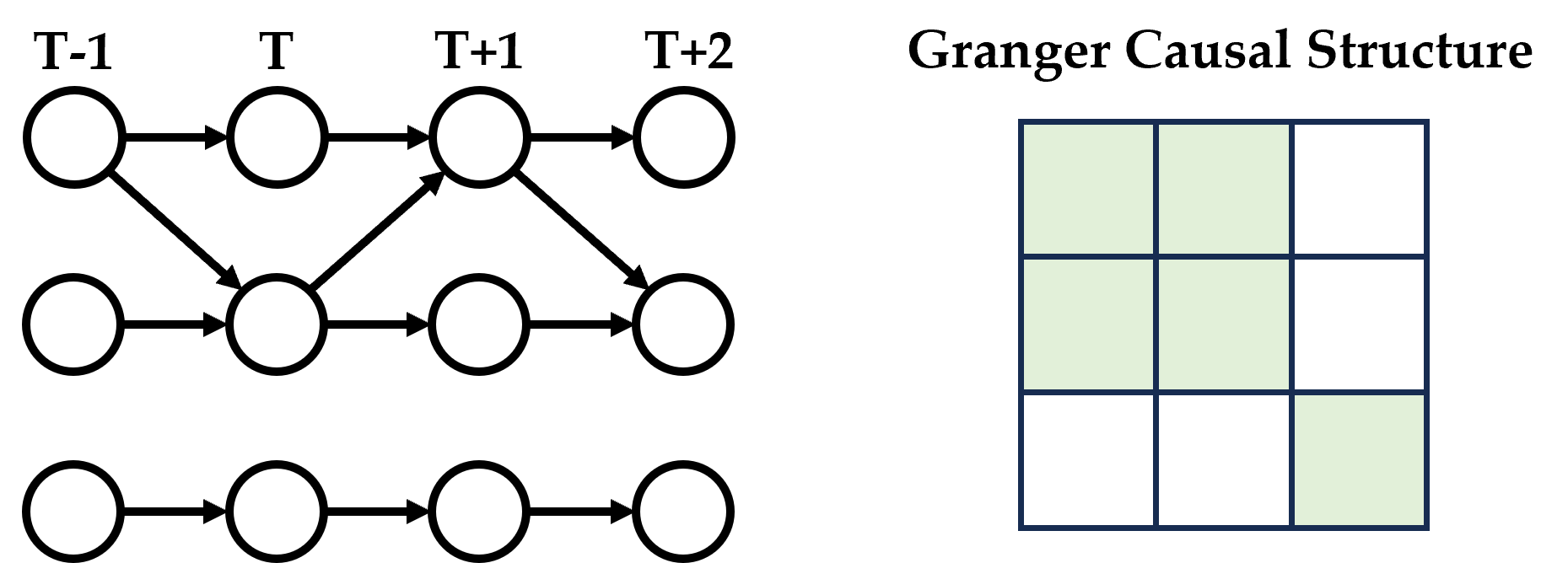}
  \caption{The complex interactions in time series data (left) lead to a Granger causal structure (right) that is not a strict DAG.}
  \label{fig:granger}
  \Description{}
\end{figure}
\noindent The identifiability of Granger causal structure for observational time series data has been established, where the parameters $\textbf{W}$ can be identified from standard results in vector autoregressive (VAR) models \cite{pamfil2020dynotears}. For linear time series interventional data, the identifiability results have been studied in \cite{chen2018two}. Specifically, the model is identifiable if each variable is influenced by a unique set of intervened variables.
\noindent In the context of non-linear interventional time series data with known interventional targets, \cite{gao2022idyno} expanded upon the $\mathcal{I}$-Markov Equivalence Class to $(\mathcal{I},\mathcal{D})$-Markov Equivalence Class for graphs within a subset of DAGs rather than all DAGs. However, addressing the challenge of identifying Granger causal structure in non-linear time series data with unknown interventional targets remains a significant and unresolved area of research. To address this issue, our initial step is to establish the negative score function for a DAG $\mathcal{G}$:
\begin{equation}
    -\mathcal{S}_{\mathcal{I}}(\mathcal{G}) := \mathbb{E}_{\textbf{X}|{\textbf{P}_{e_k}},\mathcal{G}^{*}}[\mathcal{L}_{reg}(\textbf{X})],
\end{equation}
where $\mathcal{L}_{reg}$ denotes the regularized loss minimized in Equation (\ref{eq9}), with the loss $||\cdot||_2^{2}$ being negative log-liklihood, over the time series data $\textbf{X}$, which is generated from the ground truth $\mathcal{G}^{*}$, under the interventional distribution $\textbf{P}_{e_k}$ for each $\textbf{I}_k \in \mathcal{I}$. Based on these definitions, the following theorem holds:
\begin{theorem}\label{t1}
    Let $\hat{\mathcal{G}} \in \mathcal{D}$ be a DAG and $\hat{\mathcal{I}}$ be an interventional family, which $(\hat{\mathcal{G}}, \hat{\mathcal{I}}) \in \argmax_{\mathcal{G},\mathcal{I}}\mathcal{S}(\mathcal{G},\mathcal{I})$. Under the assumption that the density models have sufficient capacity to represent the ground truth distribution, that $\mathcal{I}^{*}$-faithfulness holds, that the density models are strictly positive, that the ground truth densities $\mathbf{P}_{e_k}$ have differentiable entropy. For $\lambda_{\mathcal{G}}, \lambda_{\mathcal{I}}>0$ in Equation (\ref{eq9}) small enough, $\hat{\mathcal{G}}$ is $(\mathcal{I}^{*},\mathcal{D})$-Markov equivalent to $\mathcal{G}^{*}$ and $\hat{\mathcal{I}} = \mathcal{I}^{*}$.
\end{theorem}
\noindent The Granger causal strcture we've learned is not a strict DAG due to the intricate nature of time series data, as shown in Figure \ref{fig:granger}. However, rather than focusing directly on the Granger causal structure, our approach centers on the complex interactions within the time series data. Particularly, given the forward-in-time property, the unrolled temporally extended graph is a $\text{DAG} \in \mathbb{R}^{d \times T}$  and does not include any cyclic subgraphs, thus we omit the DAG constraint \cite{zheng2018dags} in our theoretical analysis. Establishing the identifiability of this DAG also allows us to identify the Granger causal structure. The \emph{IGC} methodology, characterized by its time-windowed approach, offers flexibility for detecting the causal relationship between any two variables $(x_{i,t}, x_{j,t'})$ in this DAG with a given time lag $p = t'-t$. If we set $\mathcal{D}$ to be the subset $\mathcal{D}_s$ of DAGs which correspond to stationary dynamics with constant-in-time conditional distributions (For detailed information and the proof of Theorem \ref{t1}, please refer to the Appendix \ref{uti}), Theorem \ref{t1} can be restated as follows:
\begin{corollary}
    Let $\hat{\mathcal{G}} \in \mathcal{D}_{s}$ be a DAG and $\hat{\mathcal{I}}$ be an interventional family. Given the same assumptions as Theorem \ref{t1}, and for $\lambda_{\mathcal{G}}, \lambda_{\mathcal{I}}$ in Equation (\ref{eq9}) small enough, $\hat{\mathcal{G}}$ is $(\mathcal{I}^{*},\mathcal{D}_{s})$-Markov equivalent to $\mathcal{G}^{*}$ and $\hat{\mathcal{I}} = \mathcal{I}^{*}$.
\end{corollary}
\noindent The Theorem \ref{t1} extends prior work \citep{brouillard2020differentiable,gao2022idyno} by showing that, under appropriate assumptions, maximizing $\mathcal{S}(\mathcal{G},\mathcal{I})$ with respect $\mathcal{G}$ and $\mathcal{I}$ recovers both the $(\mathcal{I}^{*},\mathcal{D})$-Markov equivalent class of $\mathcal{G}^{*}$ and the ground truth interventional family $\mathcal{I}^{*}$.

\begin{table*}[ht]
\centering
\resizebox{17.5cm}{!}{
\begin{tabular}{|c|c|c|c|c|c|c|c|c|c|}
\hline
\textbf{Dataset} & \textbf{Metrics} & VAR & PCMCI & NGC & eSRU & DyNoTears & GVAR & CUTS& \textbf{IGC}\\
\hline
\multirow{4}{*}{Linear (n=5)}
& Acc & $0.640 (\pm 0.080)$ & $0.800 (\pm 0.040)$ & $0.920 (\pm 0.000)$ & $0.960 (\pm 0.000)$ & $0.800 (\pm 0.040)$ &$0.960 (\pm 0.000)$ & $0.920 (\pm 0.000)$ &$\mathbf{1.000(\pm 0.000)}$\\

& AUROC & $0.650 (\pm 0.017)$ & $0.770 (\pm 0.012)$ & $0.925 (\pm 0.011)$ & $0.967 (\pm 0.008)$ & $0.740 (\pm 0.005)$ &$0.985(\pm 0.015)$ & $0.933 (\pm 0.005)$ &$\mathbf{1.000(\pm 0.000)}$\\

& F1 & $0.609(\pm 0.008)$& $0.667 (\pm 0.017)$ & $0.909 (\pm 0.000)$ & $0.952 (\pm 0.000)$& $0.725 (\pm 0.024)$ & $0.949(\pm 0.000)$ & $0.911 (\pm 0.000)$ &$\mathbf{1.000(\pm 0.000)}$\\

& SHD & $9 (\pm 2)$ &$5 (\pm 1)$ & $2 (\pm 0)$ & $1(\pm 0)$ & $5 (\pm 1)$ &$1(\pm 0)$ & $2 (\pm 0)$ &$\mathbf{0(\pm 0)}$\\
\hline

\multirow{4}{*}{Linear (n=10)}
& Acc & $0.560 (\pm 0.030)$ & $0.610 (\pm 0.030)$ & $0.820 (\pm 0.040)$ & $0.850 (\pm 0.050)$ & $0.650 (\pm 0.020)$ & $\mathbf{0.930(\pm 0.010)}$ & $0.880 (\pm 0.020)$ &$\mathbf{0.930(\pm 0.010)}$\\

& AUROC & $0.562 (\pm 0.024)$ & $0.710 (\pm 0.012)$ & $0.848 (\pm 0.010)$& $0.812 (\pm 0.008)$ & $0.524 (\pm 0.006)$ & $0.980(\pm 0.013)$ & $0.865 (\pm 0.042)$ &$\mathbf{0.989(\pm 0.018)}$\\

& F1 & $0.551 (\pm 0.029)$ & $0.456 (\pm 0.048)$ & $0.847 (\pm 0.014)$ &$0.869 (\pm 0.022)$ & $0.596 (\pm 0.032)$ &$0.912(\pm 0.014)$& $0.872 (\pm 0.012)$ &$\mathbf{0.928(\pm 0.017)}$\\

& SHD & $44 (\pm 3)$ & $39 (\pm 3)$ & $18 (\pm 4)$& $15 (\pm 5)$& $35 (\pm 2)$ & $\mathbf{7(\pm 1)}$ & $12 (\pm 2)$ &$\mathbf{7(\pm 1)}$\\
\hline

\multirow{4}{*}{Linear (n=20)}
& Acc & $0.518 (\pm 0.030)$ & $0.555 (\pm 0.030)$ & $0.815 (\pm 0.030)$ & $0.730 (\pm 0.020)$ & $0.565 (\pm 0.023)$ & $0.783 (\pm 0.040)$ & $0.838 (\pm 0.020)$  & $\mathbf{0.955(\pm 0.005)}$\\

& AUROC & $0.538 (\pm 0.035)$ & $0.545 (\pm 0.035)$ & $0.822 (\pm 0.011)$ & $0.723 (\pm 0.035)$ & $0.511 (\pm 0.005)$ & $0.854 (\pm 0.019)$ & $0.832 (\pm 0.017)$ &$\mathbf{0.973(\pm 0.006)}$\\

& F1 &$0.671 (\pm 0.012)$ & $0.351 (\pm 0.052)$  & $0.812 (\pm 0.000)$ & $0.772 (\pm 0.012)$& $0.322 (\pm 0.046)$ & $0.800 (\pm 0.038)$ & $0.816 (\pm 0.011)$ &  $\mathbf{0.955(\pm 0.002)}$\\

& SHD & $193 (\pm 6)$ & $178 (\pm 12)$ &$74 (\pm 12)$ & $108 (\pm 6)$& $174 (\pm 9)$ & $87 (\pm 16)$ & $65 (\pm 8)$ &  $\mathbf{18(\pm 2)}$\\
\hline
\hline
\textbf{Dataset} & \textbf{Metrics} & VAR & PCMCI & NGC & eSRU & DyNoTears & GVAR & CUTS& \textbf{IGC}\\

\hline
\multirow{4}{*}{Non-linear (n=5)}
& Acc & $0.458 (\pm 0.080)$ & $0.560 (\pm 0.040)$ & $0.960 (\pm 0.000)$ & $0.760 (\pm 0.040)$ & $0.800 (\pm 0.080)$ & $0.920 (\pm 0.040)$& $0.920 (\pm 0.040)$ & $\mathbf{1.000(\pm 0.000)}$\\

& AUROC & $0.517 (\pm 0.035)$ & $0.567 (\pm 0.009)$ & $0.967 (\pm 0.008)$ & $0.767 (\pm 0.018)$ & $0.740 (\pm 0.005)$ & $0.912 (\pm 0.019)$ & $0.935 (\pm 0.015)$ & $\mathbf{1.000(\pm 0.000)}$\\

& F1 & $0.563 (\pm 0.013)$ & $0.522 (\pm 0.012)$ & $0.952 (\pm 0.000)$ & $0.727 (\pm 0.006)$ & $0.725 (\pm 0.054)$ & $0.920 (\pm 0.020)$ & $0.915 (\pm 0.016)$ &  $\mathbf{1.000(\pm 0.000)}$\\

& SHD & $14 (\pm 2)$ & $11 (\pm 1)$ & $1 (\pm 0)$ & $6 (\pm 1)$ & $5 (\pm 2)$ & $2 (\pm 1)$ & $2 (\pm 1)$ &  $\mathbf{0(\pm 0)}$\\
\hline

\multirow{4}{*}{Non-linear (n=10)}
& Acc & $0.520 (\pm 0.020)$ & $0.580 (\pm 0.020)$ & $0.880 (\pm 0.020)$ & $0.710 (\pm 0.030)$ & $0.620 (\pm 0.030)$ & $0.920 (\pm 0.010)$ &$0.860 (\pm 0.030)$& $\mathbf{0.930(\pm 0.020)}$\\

& AUROC & $0.512 (\pm 0.004)$ & $0.626 (\pm 0.015)$ & $0.892 (\pm 0.009)$ & $0.709 (\pm 0.038)$ & $0.548 (\pm 0.008)$ & $0.901 (\pm 0.020)$ & $0.859 (\pm 0.031)$& $\mathbf{0.959(\pm 0.005)}$\\

& F1 & $0.658 (\pm 0.029)$ & $0.600 (\pm 0.020)$ & $0.893 (\pm 0.011)$ & $0.721 (\pm 0.012)$& $0.498 (\pm 0.042)$ & $0.913 (\pm 0.016)$ & $0.834 (\pm 0.009)$  &  $\mathbf{0.942(\pm 0.011)}$\\

& SHD & $48 (\pm 2)$ & $42 (\pm 2)$ & $12 (\pm 2)$ & $29 (\pm 3)$ & $38 (\pm 3)$ & $9 (\pm 1)$ & $14 (\pm 3)$ &  $\mathbf{7(\pm 2)}$\\
\hline

\multirow{4}{*}{Non-linear (n=20)}
& Acc & $0.508 (\pm 0.008)$ & $0.545 (\pm 0.025)$ & $0.795 (\pm 0.018)$ &  $0.647 (\pm 0.013)$ & $0.543 (\pm 0.020)$ & $0.825 (\pm 0.048)$ & $0.805 (\pm 0.020)$&$\mathbf{0.943(\pm 0.008)}$\\

& AUROC & $0.515 (\pm 0.010)$  & $0.548 (\pm 0.020)$ & $0.800 (\pm 0.014)$ & $0.641 (\pm 0.014)$ & $0.587 (\pm 0.008)$ & $0.882 (\pm 0.016)$ & $0.820 (\pm 0.035)$ &$\mathbf{0.950(\pm 0.015)}$\\

& F1 & $0.659 (\pm 0.008)$ & $0.461 (\pm 0.022)$ & $0.793 (\pm 0.020)$ & $0.714 (\pm 0.003)$ & $0.435 (\pm 0.017)$  & $0.821 (\pm 0.027)$ & $0.811 (\pm 0.004)$ & $\mathbf{0.944(\pm 0.006)}$\\

& SHD & $197 (\pm 3)$ & $182 (\pm 10)$ & $82 (\pm 7)$ & $141 (\pm 5)$ & $183 (\pm 8)$ & $70 (\pm 19)$ & $78 (\pm 8)$ & $\mathbf{23 (\pm 3)}$ \\
\hline
\end{tabular}
}
\caption{Comparative results (mean $\pm$ std.) for synthetic interventional datasets.}
\label{tab:syn}
\end{table*}

\section{Experiments}
We evaluate our proposed \textit{IGC}\footnote{\url{https://github.com/Tamuzzy/IGC}} for inferring Granger causal struture and compare them with various state-of-the-art~(SOTA) baselines across several interventional time series datasets, demonstrating the superior performance of our proposed $\textit{IGC}$ method. The competing SOTA methods for learning Granger causal structure that we benchmarked are listed as follows:
1) \textbf{VAR} (Vector AutoRegressive) \citep{granger1969investigating, granger1980testing} is a linear model used in Granger causality test. \textbf{PCMCI}~\citep{runge2019detecting} integrates conditional independence tests with optimized conditioning sets for inferring causal structure. \textbf{NGC} \cite{tank2021neural} includes the component-wise MLP and the component-wise LSTM, featuring sparse input weight layers, is proposed as an effective approach for inferring non-linear Granger causality. \textbf{eSRU} \cite{khanna2019economy} (economy Statistical Recurrent Units) are a specialized form of recurrent neural networks (RNNs) tailored to identify the network structure of non-linear Granger causal relationships. \textbf{DyNoTears} \cite{pamfil2020dynotears} is a score-based method with continuous optimization for learning causal structure. \textbf{GVAR} \cite{marcinkevivcs2020interpretable} model integrates SENNs with traditional vector autoregression for Granger causal inference. \textbf{CUTS} \citep{cheng2023cuts, cheng2024cuts+} is a neural Granger causal discovery algorithm for imputed and high dimensional data.

For evaluation purposes, we utilize the following metrics:
\textbf{Accuracy} refers to the rate at which a model correctly predicts the presence or absence of edges in the ground-truth graph. \textbf{AUROC} (Area Under the Receiver Operating Characteristic) curve is represented by the area under a curve plotting the true positive rate against the false positive rate at various thresholds. \textbf{AUPRC} (Area Under the Precision-Recall Curve) focuses on the relationship between precision and recall across different thresholds. The \textbf{F1 Score} represents the harmonic mean of precision and recall, with precision being the proportion of correctly detected edges relative to all edges predicted by the model. \textbf{SHD} (Structural Hamming Distance) denotes the count of incorrectly predicted edge states. \textbf{Recall} measures the fraction of edges in the ground-truth graph that are accurately identified by the model.
\subsection{Granger Causal Structure Learning}
Firstly, we follow the functional causal model \cite{huang2020causal} detailed in Equation (\ref{eq14}) to generate synthetic interventional time series data: 
\begin{equation}\label{eq14}
    X_{t}^{i} = \sum f_{i}(X_j) + \epsilon_{i,t}, 
\end{equation}
where $X_j  \in \textbf{PA}(X_{i,t})$ and the function $f_{i}$ can be selected from a set of functions which includes linear, cubic, tanh, and sinc functions, as well as their mixtures. The noise term $\epsilon_{i,t}$ is generated from either a uniform distribution $\mathcal{U}(-0.5,0.5)$ or a standard normal distribution $\mathcal{N}(0,1)$. 

\noindent\textbf{Linear Synthetic Interventional Time Series Data:}
In the linear setting, we generate the time series data by following these steps: 
\begin{itemize}[leftmargin=*]
    \item We constructed the Granger causal graph $\mathcal{G}$ by employing two tunable parameters: $n$ (number of nodes) and $p$ (probability of edge creation). In our experiments, we set $n$ to be 5, 10, and 20, and $p = 0.4$, we sample the weights uniformly at random from $\mathcal{U}([-0.6, -0.4] \cup [0.4, 0.6])$. 
    \item We generate the data with first autoregessive order, where data only depends on the previous time step. We generate 5, 10, and 20 sequences with 500 time steps with standard Gaussian noises.
    \item To generate the interventional time series data in $e_k$, we adopt an imperfect setting where the weights from $\textbf{PA}(X^{i}_{t;e_k})$ to $X^{i}_{t;e_k}$ are altered at a specific timestep $t=200$ by adding a random number within the range $\mathcal{U}([-0.15, 0) \cup (0, 0.15])$.
\end{itemize}
\noindent Table \ref{tab:syn} illustrates that \emph{IGC} achieves the best performance, and performs slightly better at $n=20$ than at $n=10$, which suggests that the distinction does not significantly weaken our method. 

\noindent \textbf{Non-linear Synthetic Interventional Time Series Data:}
To evaluate the efficiency of \emph{IGC} in the context of non-linear synthetic interventional time series data, we follow the same generation procedure as that used in the linear setting. However, instead of employing a linear function, we modify the underlying generation function $f_i$ in Equation (\ref{eq14}) to employ a 2-layer fully connected neural network with the Leaky ReLU activation and 0.1 negative non-linearity. The network weights are sampled uniformly from $\mathcal{U}([-0.6, -0.4] \cup [0.4, 0.6])$. To implement imperfect interventions, we add a random vector, drawn from $\mathcal{N}(0,1)$ to the network's second layer.

\begin{figure}[ht]
  \centering
  \includegraphics[width=0.95\linewidth]{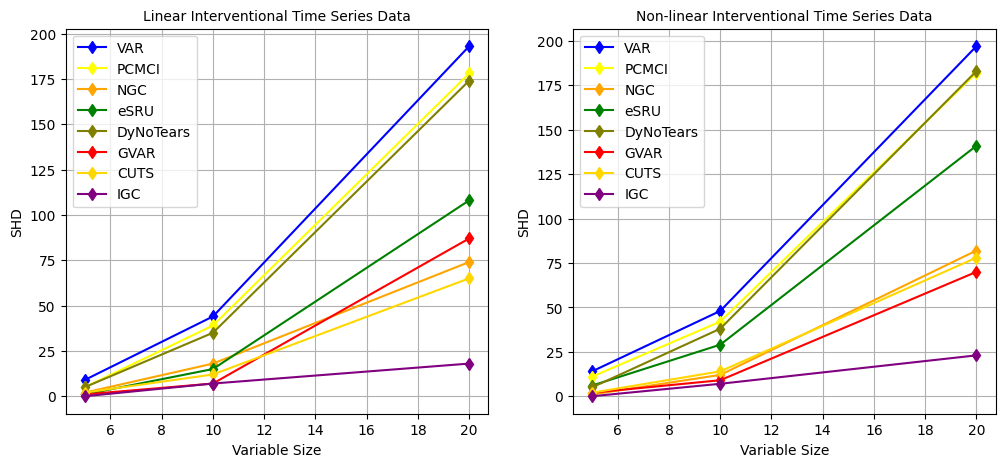}
  \captionsetup{justification=centering}
  \caption{SHD results for Linear (left) and Non-linear (right) Synthetic Interventional Time Series Data.}
  \label{fig:shd}
  \Description{}
\end{figure}

\noindent As illustrated in Table \ref{tab:syn}, we compare the performance of \emph{IGC} against other methods for $n = \{5, 10, 20\}$. The results confirm that our method consistently outperforms the others, even when the variable size is large. Figure \ref{fig:shd} presents the comparison of the results for learning Granger causal structure with varying numbers of variables in both linear and non-linear settings. From the results, we observe that the introduction of interventional data disrupts the stationary assumption underlying these models, leading to poor performance in inferring the Granger causal structure.  In contrast, our model exhibits enhanced capability in managing the non-stationarity induced by interventions, thereby achieving more accurate inference of the Granger causal structure.
\begin{table}[h]
\centering
\resizebox{8.0cm}{!}{
\begin{tabular}{ccccc}
    \toprule
    \multicolumn{5}{c}{\textbf{Lorenz-96 Model}}\\
    
    \toprule
    \multicolumn{1}{c}{\textbf{Metric/Methods}} & Acc & AUPRC & AUROC & SHD\\

    \midrule
    VAR & $0.765(\pm 0.008)$ & $0.464(\pm 0.046)$ & $0.745(\pm 0.047)$ & $94(\pm 3)$\\

    \midrule
    PCMCI & $0.720(\pm 0.020)$ & $0.724(\pm 0.007)$ & $0.788(\pm 0.033)$ & $112(\pm 8)$\\

    \midrule
    NGC & $0.653(\pm 0.028)$ & $0.956(\pm 0.016)$ & $0.979(\pm 0.016)$ & $139(\pm 11)$ \\

    \midrule
    eSRU & $0.823(\pm 0.010)$ & $0.834(\pm 0.033)$& $0.934(\pm 0.021)$ & $70(\pm 4)$\\

    \midrule
    DyNoTears & $0.785(\pm 0.013)$ & $0.779(\pm 0.035)$ & $0.811(\pm 0.015)$ & $86(\pm 5)$\\

    \midrule
    GVAR &$0.845(\pm 0.010)$ & $0.916(\pm 0.024)$& $0.970(\pm 0.009)$& $62(\pm 4)$\\

    \midrule
    CUTS & $0.755(\pm 0.023)$ & $0.785(\pm 0.015)$ & $0.876(\pm 0.017)$ & $98(\pm 9)$\\

    \midrule
    \textbf{IGC} & $\mathbf{0.925(\pm 0.008)}$ & $\mathbf{0.979(\pm 0.003)}$& $\mathbf{0.985(\pm 0.002)}$ & $\mathbf{30(\pm 2)}$\\
    
    \bottomrule
\end{tabular}}
\caption{Comparative results (mean$\pm$std.) for Lorenz-96.}
\label{tab:lorenz}
\end{table}

\noindent \textbf{Lorenz-96 Model:}
The Lorenz 96 model, a standard benchmark for Granger causal inference techniques \cite{lorenz1996predictability}, is a continuous-time dynamic system with $m$ variables, defined by non-linear differential equations:
\begin{equation}
    \frac{dx^{i}}{dt} = (x_{i+1} - x_{i-2})x_{i-1} - x_{i} + F, 
\end{equation}
where $x_0 := x_m$, $x_{-1}:=x_{m-1}$, and $x_{m+1}:=x_1$; and $F$ is a forcing constant that, in combination with $m$, controls the non-linearity of the system \citep{tank2021neural, karimi2010extensive}. We numerically simulate $m=20$ variables and $T=500$ observations under $F=40$. This choice is predicated on the understanding that a higher number of variables coupled with a higher non-linearity ($F=40$) presents a more challenging inference problem. While adhering to the experimental setup of \cite{marcinkevivcs2020interpretable}, our study introduces a more challenging setting. We manipulated the data with $m=20$ variables and $T=500$ observations under $F=40$ by altering the value of $F$ to 50 for samples when $t>250$, thus introducing an intervention in the dataset to simulate real-world complexities. From Table \ref{tab:lorenz}, we observe that our proposed \emph{IGC} achieves competitive performance even in more complex situations.

\begin{table}[ht]
\centering
\resizebox{8.0cm}{!}{
\begin{tabular}{cccccccccc}
    \toprule
    \multicolumn{10}{c}{\textbf{Tennessee Eastman Dataset}}\\
    
    \toprule
    \multicolumn{1}{c}{\textbf{Metric/Methods}} & Acc & Recall & F1 & SHD & \multicolumn{1}{c}{\textbf{Metric/Methods}} & Acc & Recall & F1 & SHD\\

    \midrule
    CORL & $0.838$ & $0.043$ & $0.071$ & $176$ & NoTears-MLP & $0.925$ & $0.036$ & $0.046$ & $82$\\

    \midrule
    DirectLiNGAM & $0.918$ & $0.046$ & $0.061$ & $89$ & PCMCI & $0.882$ & $0.094$ & $0.044$ & $129$\\

    \midrule
    FCI & $0.966$ & $0.167$ & $0.091$ & $37$ & DyNoTears & $0.928$ & $0.094$ & $0.071$ &$78$\\

    \midrule
    GES & $0.903$ & $0.040$& $0.060$ & $106$ & eSRU & $0.936$ & $0.054$ & $0.068$ & $70$\\

    \midrule
    GEOLEM & $0.890$ & $0.031$ & $0.046$ & $120$ & GVAR & $0.928$ & $0.188$ & $0.133$ & $78$\\

    \midrule
    ICALiNGAM &$0.908$ & $0.079$ & $0.116 $& $100$ & NGC & $0.852$ & $0.089$ & $\textbf{0.148}$ & $161$\\

    \midrule
    MCSL & $0.951$ & $0.080$ & $0.070$ & $53$ & CUTS & $0.922$ & $0.094$ & $0.066$ &$85$\\

    \midrule
    NoTears & $0.968$ & $0$ & $0$ & $35$ & \textbf{IGC} & $\textbf{0.968}$ & $\textbf{0.286}$& $0.103$ & $\textbf{35}$ \\
    
    \bottomrule
\end{tabular}}
\caption{Comparative results for TEP Dataset.}
\label{tab:tep}
\end{table}

\begin{figure*}[ht]
  \centering
  \includegraphics[width=0.95\linewidth]{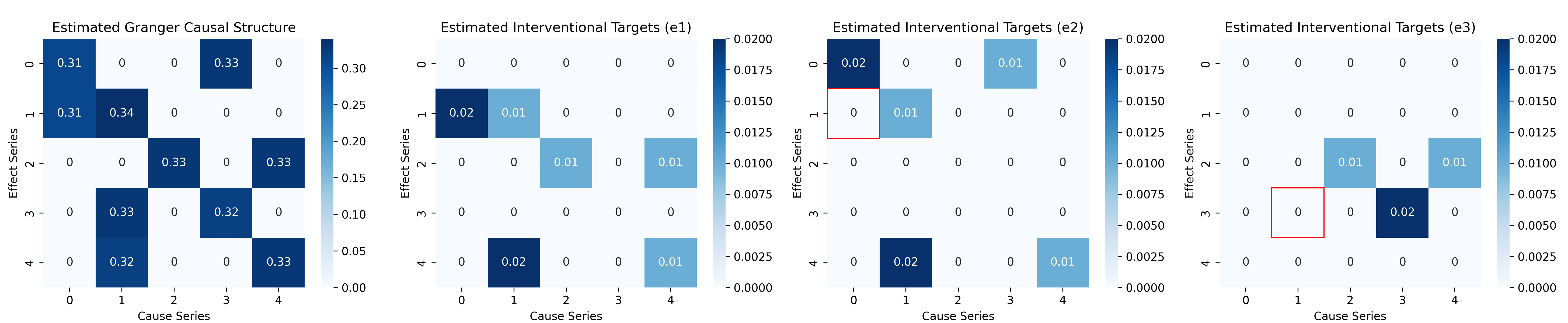}
  \caption{The estimated Granger causal structure and the estimated unknown interventional targets across three different environments, we have highlighted the discrepancies in the results using red blocks to indicate areas of disagreement.}
  \label{fig:ifr}
  \Description{}
\end{figure*}

\noindent \textbf{Tennessee Eastman Process~(TEP):}
The Tennessee Eastman Process (TEP)~\cite{downs1993plant}, serves as a widely recognized benchmark in chemical engineering research. This simulator is particularly valuable for studies in anomaly detection and root cause analysis, due to its capability to replicate process faults and the comprehensive description it offers of the entire production process. The TEP includes five principal units: a two-phase reactor, a condenser, a recycle compressor, a liquid-vapor separator, and a product stripper, involving 41 measured and 12 manipulated variables. The observational dataset is devoid of anomalies and comprises 500 observations. Within the TEP, there are 21 predefined faults, resulting in 21 distinct test datasets. Each dataset contains 960 observations, recorded at 3-minute intervals. The initial 160 observations in each dataset are anomaly-free. Starting from observation 161 and continuing to the end of the dataset, one of the 21 faults is introduced, marking a transition to conditions where the system's behavior deviates from the norm. In our research, we have utilized 22 measured variables and 11 manipulated variables. We have employed various causal structure learning methods on the observational data and integrated our propsoed \emph{IGC} approach on both observational data and interventional data with anomalies. The results, summarized in Table \ref{tab:tep}, demonstrate that our method consistently outperforms several other techniques, reinforcing its efficacy in handling interventional time series data. The higher Recall and F1-Score of our proposed method compared to NoTears can be attributed to a greater number of True Positives (TP) since the TEP dataset with a quite sparse adjacency matrix $\in \mathbb{R}^{33 \times 33}$, where only 66 elements being 1 (including 33 diagonal elements). Our method, which is based on Granger causality, leverages historical temporal information effectively. In contrast, NoTears is not specifically designed for time series data, leading to its less efficient capture of this crucial information.
\subsection{Interventional Family Recovery}
\begin{figure}[ht]
  \centering
  \includegraphics[width=0.9\linewidth]{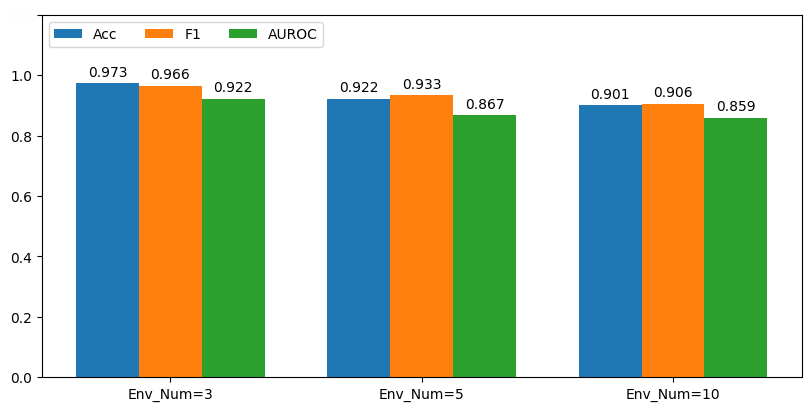}
  \captionsetup{justification=centering}
  \caption{Evaluation of the interventional target recovery.}
  \label{fig:eva}
  \Description{}
\end{figure}

So far, our focus has been on inferring Granger causal structure, without addressing the issue of interventional family recovery, which is crucial for a deeper understanding of various time series analysis tasks. Although there have been some experiments targeting known interventions, to the best of our knowledge, this study is the first to delve into the recovery of unknown interventional targets from interventional time series data. This approach not only enhances our understanding of the underlying processes but also clarifies the interaction between the Granger causal structure learning and the recovery of unknown targets. To bridge the identified gap, we assessed the model's capability in accurately identifying unknown interventional targets within synthetic datasets. We first formulate the problem as follows:
\begin{problem}
    Consider a Granger causal graph $\mathcal{G} \in \mathbb{R}^{d \times d}$, with the assumption that the time series generation follows Equation (\ref{eq14}). We introduce \textit{interventional family}, denoted as $\mathcal{I} := \{\textbf{I}_1,\ldots,\textbf{I}_k\}$, where each $\textbf{I}_k \in \mathbb{R}^{d \times d}$. This implies that, after $t$ time steps, some specific edges in the Granger causal graph $\mathcal{G}$ are chosen as interventional targets in $k$-th intervention. The problem is to recover these interventional targets based solely on the observed intervened or non-intervened time series data from several environments, without access to the knowledge of non-intervened time series data.
\end{problem}
\noindent Figure \ref{fig:ifr} illustrated that the Granger causal structure, generated from the causal network, is stable across multiple environments, despite changes in causal strength following the interventions. Regarding the intervention networks, the results highlight the effects of interventions. We used red blocks to indicate areas of disagreement in Figure \ref{fig:ifr} and we attribute these discrepancies to instances where the intervention strength was insufficient or below certain thresholds. We also evaluated the task of recovering interventional targets on synthetic interventional time series data from several distinct environments, as illustrated in Figure \ref{fig:eva}. We found that our method prioritizes environments experiencing high-intensity interventions, potentially overlooking those with milder interventions. Thus, it is important to set thresholds based on the specific application to determine whether the environment is being intervened. It is also worth noting that the number of interventions should be less than a threshold, which can be described as: $|\mathcal{I}| \leq n$.
\section{Conclusion}
In this study, we have investigated the Granger causal structure learning task, incorporating heterogeneous interventional time series data. To address the issue of identifying Granger non-causality in interventional time series data with unknown targets, we have introduced a novel condition that ensures the recovery of these unknown targets and the accurate identification of the true causal structure within the $(\mathcal{I},\mathcal{D})$-Markov Equivalence Class. We solved the identifiability issues for accurately determining causal relationships in Granger causality. Our theoretical analysis is supported by empirical results, demonstrating that our proposed Interventional Granger Causal~(\emph{IGC}) structure learning method outperforms existing methodologies in both synthetic and real-world datasets, even in the absence of interventional target information. Potential avenues for future research include applying our method to a broader spectrum of time series applications, which includes detecting anomalies and root cause analysis within time series data.
\section*{Acknowledgements}
This work was supported in part by the U.S. National Science Foundation~(NSF) grants SHF-2215573, and by the U.S. Department of Engergy~(DOE) Office of Science, Advanced Scientific Computing Research (ASCR) under Awards B\&R\# KJ0403010/FWP\#CC132 and FWP\#CC138. Portions of this research were conducted with the advanced computing resources provided by Texas A\&M High Performance Research Computing.

\bibliographystyle{ACM-Reference-Format}
\bibliography{sample-base}

\appendix
\section{Appendix}
\subsection{Proximal Gradient Descent Updates}
\label{pgd}
As for Equation (\ref{eq9}), we establish the following definitions:
\begin{equation}
\begin{aligned}
    g(\phi) := \sum_{k=1}^{n}\sum_{t=2}^{T}\sum_{i=1}^{d}\sum_{j=1}^{d}|| x_{i,t}^{e_k} - \textbf{F}_{ij}\big(x_{j,t-1:1}^{e_k}; (\textbf{W}_{:,j,e_{0}}^{i}+\textbf{W}_{:,j,e_{k}}^{i}) \big) ||_2^2.
\end{aligned}
\end{equation}

\begin{equation}
\begin{aligned}
    h(\phi) :&= (1-\alpha)\lambda \sum_{i=1}^{d}\sum_{j=1}^{d} ||(\textbf{W}_{:,j,e_{0}}^{i}, \textbf{W}_{:,j,e_{1}}^{i},...,\textbf{W}_{:,j,e_{k}}^{i})||_{2}\\
    &+ \alpha \lambda \sum_{i=1}^{d}\sum_{j=1}^{d}\sum_{k=1}^{n}||\textbf{W}_{:,j,e_{k}}^{i}||_2.
\end{aligned}
\end{equation}   
A proximal mapping for the function $h(\phi)$ can be defined as follows:
\begin{equation}
\begin{aligned}
    \textbf{prox}_h(\textbf{u}) &= \argmin_{\textbf{z}}{\frac{1}{2}||\textbf{z}-\textbf{u}||_2^2} + (1-\alpha) \lambda||\textbf{z}||_2\\
    &+ \alpha\lambda \sum^{n}_{i=1}||z_i||_2.
\end{aligned}
\end{equation}
For $k = 0, 1, \ldots, n$, the updating steps at the $m$-th iteration are represented as:
\begin{equation}
\begin{aligned}
    \textbf{W}_{:,j,e_k}^{i(m)} = \textbf{prox}_{h,t_k}(\textbf{W}_{:,j,e_k}^{i(m-1)} - t_k\nabla_{\textbf{W}_{:,j,e_k}^{i}} g(\phi^{(m-1)})).
\end{aligned}
\end{equation}
Thus, $\textbf{u} = \{u_0, u_1, \ldots, u_n\}$ is a vector, and it is defined as:
\begin{equation}
\begin{aligned}
    u_k =\textbf{W}_{:,j,e_k}^{i(m-1)} - t_k\nabla_{\textbf{W}_{:,j,e_k}^{i}} g(\phi^{(m-1)}),
\end{aligned}
\end{equation}
and
$$
\textbf{W}_{:,j,e_k}^{i(m)} = \textbf{prox}_{h}(u_k).
$$
First, let's examine the scenario when $\textbf{z} = 0$. According to the Karush-Kuhn-Tucker (KKT) conditions, we obtain the following:
\begin{equation}\label{eq20}
    0 \in 
    \begin{bmatrix} z_0\\ z_1\\...\\z_n\end{bmatrix} - \begin{bmatrix} u_0\\ u_1\\...\\u_n\end{bmatrix} + 
    (1-\alpha)\lambda 
    \begin{bmatrix} \frac{z_0}{||\textbf{z}||_2}\\ \frac{z_1}{||\textbf{z}||_2}\\...\\\frac{z_n}{||\textbf{z}||_2}\end{bmatrix} +
    \alpha\lambda 
    \begin{bmatrix} 0\\ \frac{z_1}{||z_1||_2}\\...\\\frac{z_n}{||z_n||_2}\end{bmatrix}.
\end{equation}
One could set $\textbf{z}=0$, while Equation (\ref{eq21}) holds:
\begin{equation}\label{eq21}
    \begin{bmatrix} u_0\\ u_1\\...\\u_n\end{bmatrix} -
    \alpha\lambda 
    \begin{bmatrix} 0\\ \frac{z_1}{||z_1||_2}\\...\\\frac{z_n}{||z_n||_2}\end{bmatrix} =
    (1-\alpha)\lambda 
    \begin{bmatrix} \frac{z_0}{||\textbf{z}||_2}\\ \frac{z_1}{||\textbf{z}||_2}\\...\\ \frac{z_n}{||\textbf{z}||_2}\end{bmatrix}.
\end{equation}
Identifying edge cases for $\textbf{u}$ is straightforward as it involves an element-wise comparison between $\textbf{u}$ and $\textbf{z}$. Additionally, it is worth noting that $||\textbf{z}||_2 \leq 1$, leading to the following considerations:
\begin{equation}
    \textbf{z} = 0 \Leftrightarrow
    ||\textbf{u} - \alpha\lambda \begin{bmatrix} 0\\ \frac{u_1}{||u_1||_2}\\...\\ \frac{u_n}{||u_n||_2} \end{bmatrix}||_2 \leq (1-\alpha)\lambda.
\end{equation}
In the case $\textbf{z} \neq 0$, Equation (\ref{eq20}) suggests:
\begin{equation}\label{eq23}
    \begin{bmatrix} u_0\\ u_1\\...\\u_n\end{bmatrix} - \alpha\lambda 
    \begin{bmatrix} 0\\ \frac{z_1}{||z_1||_2}\\...\\\frac{z_n}{||z_n||_2}\end{bmatrix} =
    (1-\alpha)\lambda 
    \begin{bmatrix} \frac{z_0}{||\textbf{z}||_2}\\ \frac{z_1}{||\textbf{z}||_2}\\...\\\frac{z_n}{||\textbf{z}||_2}\end{bmatrix} +
    \begin{bmatrix} z_0\\ z_1\\...\\z_n\end{bmatrix}. 
\end{equation}
When considering elements in $\textbf{z}$ that are non-zero, their sign aligns with the corresponding element in $\textbf{u}$. Now, let's establish the following definition:
\begin{equation}
    S_{\alpha\lambda}(\textbf{u}) = \begin{bmatrix} u_0\\ u_1 - \alpha\lambda \frac{u_1}{||u_1||_2}\\...\\ u_n - \alpha\lambda \frac{u_n}{||u_n||_2} \end{bmatrix}.
\end{equation}
An alternative representation of Equation (\ref{eq23}) is achieved by transforming it into:
\begin{equation}\label{eq28}
    S_{\alpha\lambda}(\textbf{u}) =
    (1 + \frac{(1-\alpha)\lambda}{||\textbf{z}||_2})
    \begin{bmatrix} z_0\\ z_1\\...\\z_n\end{bmatrix}. 
\end{equation}
If we apply the L2 norm to both sides as follows:
\begin{equation}\label{eq29}
\begin{aligned}
    ||S_{\alpha\lambda}(\textbf{u})||_2 &=
    (1 + \frac{(1-\alpha)\lambda}{||\textbf{z}||_2}) \cdot
    ||\textbf{z}||_2\\
    \Rightarrow ||\textbf{z}||_2 &= ||S_{\alpha\lambda}(\textbf{u})||_2 - (1-\alpha)\lambda.
\end{aligned}
\end{equation}
Upon substituting Equation (\ref{eq29}) into Equation (\ref{eq28}), we obtain:
\begin{equation}\label{eq12}
    \textbf{z} = (1 - \frac{(1-\alpha)\lambda}{||S_{\alpha\lambda}(\textbf{u})||_2}) \cdot S_{\alpha\lambda}(\textbf{u})
\end{equation}
To summarize:
\begin{equation}
\begin{aligned}
\textbf{prox}_h(\textbf{u}) &=
\begin{cases}
0 & \textbf{if} \hspace{0.2cm} ||S_{\alpha\lambda}||_2 \leq (1-\alpha)\lambda\\
(1 - \frac{(1-\alpha)\lambda}{||S_{\alpha\lambda}||_2}) \cdot S_{\alpha\lambda}(\textbf{u}) & \textbf{if} \hspace{0.2cm} ||S_{\alpha\lambda}||_2 > (1-\alpha)\lambda
\end{cases}\\
&= (1 - \frac{(1-\alpha)\lambda}{\max \big (||S_{\alpha\lambda}||_2, (1-\alpha)\lambda \big )}) \cdot S_{\alpha\lambda}(\textbf{u})
\end{aligned}
\end{equation}

\subsection{Discussion and the Proof of Theorem 5.1.}
\label{uti}
The identifiability condition for the unrolled, temporally extended $\text{DAG} \in \mathbb{R}^{d \times T}$, which includes all variables across all time steps, has been established in the work~\cite{gao2022idyno}. Specifically, it assumes that the edges within the graph $\mathcal{G}^{*}$ remain constant over time. Furthermore, it assumes that for any given window $X \in \mathbb{R}^{d \times w}$, where $w$ represents the window's width, the distribution $\mathbf{P}_X$ over the variables within this window stays invariant across different timesteps. This implies that the conditional distribution $\mathbf{P}(x_{i,t}|\mathbf{PA}(x_{i,t}))$ for any variable $x_i$ is independent of the time index $t$. The subset of all DAGs that can be segmented in this manner into a directed sequence of a repeating subgraph or window is defined as $\mathcal{D}_s$. This segmentation is based on the fact that repetition of the same conditional distributions and edges over time corresponds to stationary or fixed dynamics. The \emph{IGC} methodology, notable for its time-windowed framework, provides the flexibility to detect causal relationship in $\mathbf{W}_{:,j,t',e_0}^{i,t}$ between any pair of variables $(x_{i,t}, x_{j,t'})$ with a given time lag $p = t'-t$ across any DAG or within any time window. When the assumptions outlined in Theorem \ref{t1}. holds, Theorem 1 from \cite{brouillard2020differentiable} becomes applicable, ensuring that our learned graph $\hat{\mathcal{G}}$ is $\mathcal{I}$-Markov equivalent to $\mathcal{G}^{*}$. Additionally, given that $\hat{\mathcal{G}} \in \mathcal{D}$, by invoking the Theorem 3.2 from \cite{gao2022idyno}, $\hat{\mathcal{G}}$ is $(\mathcal{I},\mathcal{D})$-Markov equivalent to $\mathcal{G}^{*}$. Since the true Granger causal structure is originally derived from $\mathcal{G}^{*}$, by establishing $\hat{\mathcal{G}}$ is $(\mathcal{I},\mathcal{D})$-Markov equivalent to $\mathcal{G}^{*}$, we have addressed the challenges related to the identifiability in Granger causality and mitigate the concerns associated with accurately determining causal relationships within the framework of Granger causality.
\noindent Theorem 1 from \cite{brouillard2020differentiable} and Theorem 3.2 from \cite{gao2022idyno} operate under the implicit assumption that, for each intervention $k$, the ground truth interventional target $\textbf{I}_{e_k}^{*}$ is precisely known. This assumption, however, does not always available in real-world scenarios. To address this discrepancy, we propose an extension to Theorem 3.2 from \cite{gao2022idyno} that accommodates unknown interventional targets. In this context, as our proposed \emph{IGC} method, the interventional targets $\mathcal{I}$ are learned in a manner similarly to how the graph $\mathcal{G}$ is determined. We are now ready to prove our Theorem \ref{t1}.

\begin{proof}
Leveraging Theorem 2 from \cite{brouillard2020differentiable}, we address scenarios where $\mathcal{I} \neq \mathcal{I}^{*}$. The core concept of the proof is that $\mathcal{S}(\mathcal{G}^{*}, \mathcal{I}^{*}) > \mathcal{S}(\mathcal{G}, \mathcal{I})$ whenever $\mathcal{G} \notin (\mathcal{I}^{*},\mathcal{D})$-MEC($\mathcal{G}^{*}$) or when $\mathcal{I} \neq \mathcal{I}^{*}$. For the sake of clarity, we define:
\begin{equation}
    \eta(\mathcal{G}, \mathcal{I}) := \inf_{\phi}\sum_{k \in [K]}D_{KL}(P_{e_k} || F^{e_k}_{\mathcal{G}\mathcal{I}\phi}).
\end{equation}

\begin{lemma}\label{A1}
    Let $i \in V$ and $A \subset V \backslash \{i\}$, if $(p^1,p^2) \notin \mathcal{Z}(i,A)$ and both $p^1$ and $p^2$ are strictly positive, then:
    \begin{equation}
    \begin{aligned}
    \inf_{(f^1,f^2) \in \mathcal{Z}(i,A))} D_{KL}(p^1 || f^1) + D_{KL}(p^2 || f^2) > 0.
    \end{aligned}
    \end{equation}
\end{lemma}

\begin{case}\label{c1}
Let $\mathbb{I}$ represent the set of all intervention sets $\mathcal{I}$ for which there is at least one intervention $k_0 \in [K]$ and one variable $i \in [d]$ such that $i$ is included in the true intervention set $\textbf{I}_{k_0}^{*}$ but is not included in $\textbf{I}_{k_0}$. Assuming $\mathcal{I} \in \mathbb{I}$ and considering $\mathcal{G}$ as an arbitrary DAG, the principle of $\mathcal{I}^{*}$-faithfulness implies that
\begin{equation}
    P_{e_0}(x_i|\textbf{PA}(x_i)) \neq P_{e_{k_0}}(x_i|\textbf{PA}(x_i)).
\end{equation}
It also means $(P_{e_0}, P_{e_{k_0}}) \notin \mathcal{Z}(i, \textbf{PA}(i))$, where,
\begin{equation}
    \mathcal{Z}(i,A):=\{(f^{1}, f^{2}) | f^{1}(x_i|x_A) = f^{2}(x_i|x_A) \hspace{0.1cm} \text{and} \hspace{0.1cm} f^{1}, f^{2} > 0\}.
\end{equation}
Given that $i \notin \textbf{I}_{k_0}$, it follows from the definition provided in Equation (\ref{eq7}) that, for all values of $\phi$,
\begin{equation}
\begin{aligned}
    F_{\mathcal{G}\mathcal{I}\phi}^{e_0}(x_{i}|\textbf{PA}(x_{i})) &= F_{\mathcal{G}\mathcal{I}\phi}^{e_{k_0}}(x_{i}|\textbf{PA}(x_{i})) \\ 
    \textbf{i.e.} (F_{\mathcal{G}\mathcal{I}\phi}^{e_0}, F_{\mathcal{G}\mathcal{I}\phi}^{e_{k_0}}) &\in \mathcal{Z}(i, \textbf{PA}(i)).
\end{aligned}
\end{equation}

\noindent The following holds and for all $\phi$ we have $(F^{e_0}, F^{e_{k_0}}) \in \mathcal{Z}(i, \textbf{PA}(i))$ due to Lemma \ref{A1}:
\begin{equation}
\begin{aligned}
    \eta (\mathcal{G},\mathcal{I}) &\geq \inf_{\phi} D_{KL}(P_{e_0}||F_{\mathcal{G}\mathcal{I}\phi}^{e_0}) + D_{KL}(P_{e_{k_0}}||F_{\mathcal{G}\mathcal{I}\phi}^{e_{k_0}})\\
    &\geq \inf_{(F^{e_0},F^{e_{k_0}}) \in \mathcal{Z}(i,\textbf{PA}(i))} D_{KL}(P_{e_0}||F^{e_0}) + D_{KL}(P_{e_{k_0}}||F^{e_{k_0}})\\
    & > 0.
\end{aligned}
\end{equation}

\noindent For $\min\{|\mathcal{G}| - |\mathcal{G}^{*}|, |\mathcal{I}| - |\mathcal{I}^{*}|\} \geq 0$, then $\mathcal{S}(\mathcal{G}^{*}, \mathcal{I}^{*}) > \mathcal{S}(\mathcal{G}, \mathcal{I})$. Let us define $\mathbb{S} := \{(\mathcal{G}, \mathcal{I}) \in DAG \times \mathbb{I} | \min\{|\mathcal{G}| - |\mathcal{G}^{*}|, |\mathcal{I}| - |\mathcal{I}^{*}|\} < 0\}$. To prove that $\mathcal{S}(\mathcal{G}^{*}, \mathcal{I}^{*}) - \mathcal{S}(\mathcal{G}, \mathcal{I}) > 0$ for all $(\mathcal{G}, \mathcal{I}) \in \mathbb{S}$, we need to choose $\lambda_{\mathcal{G}}, \lambda_{\mathcal{I}} > 0$ small enough. Choosing $\lambda_{\mathcal{G}} + \lambda_{\mathcal{I}} < \min_{(\mathcal{G}, \mathcal{I}) \in \mathbb{S}} \frac{\eta(\mathcal{G}, \mathcal{I})}{-\min\{|\mathcal{G}| - |\mathcal{G}^{*}| , |\mathcal{I}| - |\mathcal{I}^{*}|\}}$ since:
\begin{equation}
\begin{aligned}
    &\lambda_{\mathcal{G}} + \lambda_{\mathcal{I}} < \min_{(\mathcal{G}, \mathcal{I}) \in \mathbb{S}} \frac{\eta(\mathcal{G}, \mathcal{I})}{-\min\{|\mathcal{G}| - |\mathcal{G}^{*}| , |\mathcal{I}| - |\mathcal{I}^{*}|\}} \\
    &\Leftrightarrow \lambda_{\mathcal{G}} + \lambda_{\mathcal{I}} < \frac{\eta(\mathcal{G}, \mathcal{I})}{-\min\{|\mathcal{G}| - |\mathcal{G}^{*}| , |\mathcal{I}| - |\mathcal{I}^{*}|\}};  \forall (\mathcal{G}, \mathcal{I}) \in \mathbb{S} \\
    &\Leftrightarrow -(\lambda_{\mathcal{G}} + \lambda_{\mathcal{I}}) \min\{|\mathcal{G}| - |\mathcal{G}^{*}| , |\mathcal{I}| - |\mathcal{I}^{*}|\} < \eta (\mathcal{G},\mathcal{I})\\
    &\Leftrightarrow 0 < \eta (\mathcal{G},\mathcal{I}) + (\lambda_{\mathcal{G}} + \lambda_{\mathcal{I}}) \min\{|\mathcal{G}| - |\mathcal{G}^{*}| , |\mathcal{I}| - |\mathcal{I}^{*}|\},
\end{aligned}
\end{equation}
then we have: 
\begin{equation}
\begin{aligned}
    0 &< \eta (\mathcal{G},\mathcal{I}) + (\lambda_{\mathcal{G}} + \lambda_{\mathcal{I}}) \min\{|\mathcal{G}| - |\mathcal{G}^{*}| , |\mathcal{I}| - |\mathcal{I}^{*}|\}; \forall (\mathcal{G}, \mathcal{I}) \in \mathbb{S}\\
    & \leq \eta (\mathcal{G},\mathcal{I}) + \lambda_{\mathcal{G}}(|\mathcal{G}| - |\mathcal{G}^{*}|) + \lambda_{\mathcal{I}}(|\mathcal{I}| - |\mathcal{I}^{*}|)\\
    & = \mathcal{S}(\mathcal{G}^{*}, \mathcal{I}^{*}) - \mathcal{S}(\mathcal{G}, \mathcal{I}).
\end{aligned}
\end{equation}
\end{case}
\noindent From this point forward, we can assume that $\textbf{I}_{k}^{*} \subset \textbf{I}_{k}$ for all $k \in [K]$, and this assumption is valid because any deviation from this condition would fall under Case \ref{c1}.

\begin{lemma}\label{A2}
    Given the assumptions outlined in Theorem \ref{t1}:
    \begin{equation}
    \begin{aligned}
        \mathcal{S}(\mathcal{G}^{*},\mathcal{I}^{*}) &= \inf_{\phi}\sum_{k \in [K]} D_{KL}(P_{e_k}||F^{e_k}_{\mathcal{G}\mathcal{I}\phi})\\
        &+ \lambda_{\mathcal{G}}(|\mathcal{G}| - |\mathcal{G}^{*}|) + \lambda_{\mathcal{I}}(|\mathcal{I}| - |\mathcal{I}^{*}|).
    \end{aligned}
    \end{equation}
\end{lemma}

\begin{case}\label{c2}
Let $\overline{\mathbb{I}}:=\{\textbf{I}|\textbf{I}_k^{*} \subset \textbf{I}_k \forall k\} \text{and} [\exists (k_0,i) $\text{ s.t. }$ i \in \textbf{I}_{k_0} $\text{ and }$ i \notin \textbf{I}_{k_0}^{*}]\}$. Given that $\mathcal{I} \in \overline{\mathbb{I}}$ and $\mathcal{G}$ is a DAG, it becomes evident that $|\mathcal{I}| > |\mathcal{I}^{*}|$.
If $|\mathcal{G}| \geq |\mathcal{G}^{*}|$, then $\mathcal{S}(\mathcal{G}^{*}, \mathcal{I}^{*}) - \mathcal{S}(\mathcal{G}, \mathcal{I}) > 0$ by Lemma \ref{A2}. Define a set $\overline{\mathbb{S}} := \{(\mathcal{G}, \mathcal{I}) \in DAG \times \overline{\mathbb{I}} | |\mathcal{G}| < |\mathcal{G}^{*}|\}$. To prove that $\mathcal{S}(\mathcal{G}^{*}, \mathcal{I}^{*}) - \mathcal{S}(\mathcal{G}, \mathcal{I}) > 0$ for all $(\mathcal{G}, \mathcal{I}) \in \overline{\mathbb{S}}$, we need to choose $\lambda_\mathcal{G}$ small enough. Choosing $\lambda_{\mathcal{G}} < \min_{(\mathcal{G},\mathcal{I}) \in \overline{\mathbb{S}}} \frac{\eta(\mathcal{G}, \mathcal{I}) + \lambda_{\mathcal{I}}(|\mathcal{I}| - |\mathcal{I}^{*}|)}{|\mathcal{G}| - |\mathcal{G}^{*}|}$ since:
\begin{equation}
\begin{aligned}
    \lambda_{\mathcal{G}} &< \frac{\eta(\mathcal{G}, \mathcal{I}) + \lambda_{\mathcal{I}}(|\mathcal{I}| - |\mathcal{I}^{*}|)}{|\mathcal{G}| - |\mathcal{G}^{*}|}; \forall (\mathcal{G}, \mathcal{I}) \in \overline{\mathbb{S}}\\
    & \Leftrightarrow \lambda_{\mathcal{G}} (|\mathcal{G}| - |\mathcal{G}^{*}|) < \eta(\mathcal{G}, \mathcal{I}) + \lambda_{\mathcal{I}}(|\mathcal{I}| - |\mathcal{I}^{*}|)\\
    & \Leftrightarrow 0 < \eta (\mathcal{G}, \mathcal{I}) + \lambda_{\mathcal{G}}(|\mathcal{G}| - |\mathcal{G}^{*}|) + \lambda_{\mathcal{I}}(|\mathcal{I}| - |\mathcal{I}^{*}|),
\end{aligned}
\end{equation}
then we have:
\begin{equation}
\begin{aligned}
    0 &< \eta (\mathcal{G}, \mathcal{I}) + \lambda_{\mathcal{G}}(|\mathcal{G}| - |\mathcal{G}^{*}|) + \lambda_{\mathcal{I}}(|\mathcal{I}| - |\mathcal{I}^{*}|); \forall (\mathcal{G}, \mathcal{I}) \in \overline{\mathbb{S}}\\
    & = \mathcal{S}(\mathcal{G}^{*}, \mathcal{I}^{*}) - \mathcal{S}(\mathcal{G}, \mathcal{I}).
\end{aligned}
\end{equation}
\end{case}

\noindent In the scenarios described by Case \ref{c1} and Case \ref{c2}, all instances where $\mathcal{I} \neq \mathcal{I}^{*}$ are accounted for. Consequently, this leads to the conclusion that $\mathcal{I}$ must be equal to $\mathcal{I}^{*}$. By noting that $\mathcal{S}(\mathcal{G}^{*}, \mathcal{I}^{*}) - \mathcal{S}(\mathcal{G}, \mathcal{I}) = \mathcal{S}_{\mathcal{I}^{*}}(\mathcal{G}^{*}) - \mathcal{S}_{\mathcal{I}^{*}}(\mathcal{G})$, we can employ the same steps as \cite{gao2022idyno} to prove that $\hat{\mathcal{G}} \in (\mathcal{I}^{*},\mathcal{D})-\text{MEC}(\mathcal{G}^{*})$.
\end{proof}
\end{document}